**Large Language Models as symbolic DNA of cultural dynamics**

Parham Pourdavood[1,2], Michael Jacob[1,2,3], Terrence Deacon[1,4]

Affiliations:

1. Human Energy, 21 Orinda Way, Suite C 208 Orinda, CA 94563, United States

2. Mental Health Service, San Francisco VA Medical Center, 4150 Clement St, San Francisco, CA 94121 United States

3. Department of Psychiatry and Weill Institute for Neurosciences, University of California, San Francisco, 505 Parnassus Ave, San Francisco, CA 94143 United States

4. Department of Anthropology, University of California, Berkeley, 110 Sproul Hall, Berkeley, CA, 94720 United States

Corresponding Author:

Name: Parham Pourdavood

Email: Parham.Pourdavood@gmail.com

**Abstract**

This paper proposes a novel conceptualization of Large Language Models (LLMs) as externalized informational substrates that function analogously to DNA for human cultural dynamics. Rather than viewing LLMs as either autonomous intelligence or mere programmed mimicry, we argue they serve a broader role as repositories that preserve compressed patterns of human symbolic expression—"fossils" of meaningful dynamics that retain relational residues without their original living contexts. Crucially, these compressed patterns only become meaningful through human reinterpretation, creating a recursive feedback loop where they can be recombined and cycle back to ultimately catalyze human creative processes. Through analysis of four universal features—compression, decompression, externalization, and recursion—we demonstrate that just as DNA emerged as a compressed and externalized medium for preserving useful cellular dynamics without containing explicit reference to goal-directed physical processes, LLMs preserve useful regularities of human culture without containing understanding of embodied human experience. Therefore, we argue that LLMs' significance lies not in rivaling human intelligence, but in providing humanity a tool for self-reflection and playful hypothesis-generation in a low-stakes, simulated environment. This framework positions LLMs as tools for cultural evolvability, enabling humanity to generate novel hypotheses about itself while maintaining the human interpretation necessary to ground these hypotheses in ongoing human aesthetics and norms.

**1 Introduction**

The true nature of Artificial Intelligence (AI) and its role in human society remain subjects of contentious and ongoing debate. Although the recent wave of AI models, known as Large Language Models (LLMs), are seamlessly surpassing the Turing Test, this milestone has been overshadowed by their rapid commercialization and the profound ways they are already reshaping society. The pursuit of Artificial General Intelligence (AGI)—commonly defined as human-level intelligence—is touted as the next major milestone. Yet whether the continued progress within the current framework could ever lead to agency and meaning at the scale of AI itself remains an open and contested question.

Critics argue that current LLMs operate through algorithmic mimicry, that is simulating intelligent behavior without embodying the principles behind it (Jaeger, 2024; Jaeger et al., 2024). Artificial Neural Networks—the main framework behind LLMs—operate on behaviorist assumptions: a framework that focuses exclusively on observable input-output patterns while treating internal states as part of a "black box" to be optimized (Brooks, 1991; Sutton & Barto, 2015). This does not mean LLMs do not have sophisticated engineering, but their structure is designed to optimize internal states based on input-output feedback loops. Even though the logic behind behaviorism is likely one of the key principles supporting an intelligent system, it likely is not sufficient for intelligence and is not what enables agency and intelligence in the first place (Dreyfus, 1992; Searle, 1980). Furthermore, it would be naive to consider outward behavior of intelligence as having acquired intelligence or sentience since a good simulation can be powerful and convincing.

To address such issues, alternative approaches grounded in organismal intelligence are emerging to instead explain the principles behind intelligence through intrinsic and goal-directed models of the body and its relationship to the environment (Deacon, 2012; Jacob, 2023; Jaeger et al., 2024; Levin, 2019; Roli et al., 2022; Varela et al., 1993; Watson, 2024). Regardless, once the principles behind intelligence are fully understood—wherever they may be—the possibility of implementing general intelligence in artificial form remains theoretically possible.

Until such a feat is achieved, and rather than dismissing the dominant computational paradigm entirely, we examine what role these mimicry-based systems might serve for cultural dynamics. This perspective aligns with emerging scholarship that positions AI systems within the broader context of human cultural processes rather than as isolated technological entities. In this perspective, AI is not an autonomous intelligence but a technology deeply embedded within human cultural processes.

For example, Jaron Lanier describes LLMs as "an innovative form of social collaboration" and responsible for "illuminating previously hidden concordances between human creations"—a human correlation finder. Lanier argues that LLMs illuminate hidden patterns across human creations, functioning as statistical mash up of human contributions (Lanier, 2023). Similarly, Ted Chiang characterizes LLMs as performing lossy compression of human knowledge, extracting statistical regularities that enable generative output through gap-filling, where "hallucinations" emerge as compression artifacts rather than malfunctions (Chiang, 2023). One of us (TD) has extended this view by describing LLMs as "fossils of meaning and use"—preserving relational residues of human interactions, without access to the original contexts, referents, or embodied experiences that gave those interactions meaning (Deacon, 2023). Like geological fossils, these systems capture the structural contours of meaningful processes in compressed, reconfigurable form, yet remain detached from the spatiotemporal and living dynamics that produced them (Nosta, 2025) .

More concretely, we examine whether LLMs are implementing a process for cultural dynamics which mirrors a common process in biology: externalization of useful embodied dynamics in a compressed medium in order to reliably preserve and catalyze them, which is similar to the function of DNA in cells or memory in the brain. Even though the compressed forms within LLMs contain useful contextual clues of human-generated data, by themselves they do not carry intelligence or meaning; they need to be decompressed by the already embodied and meaningful dynamics of humans in order to acquire renewed meaning and to become grounded again. In this light, LLMs can act as a flexible medium that allows humanity to access and explore combinations of its useful symbolic patterns, which may afford humanity as a whole a form of self-reflective innovation.

Since our argument positions LLMs as extensions of human symbolic processes, we begin by clarifying our conception of symbols and examining their fundamental characteristics.

**2 What Are Symbols?**

Our argument rests on an important insight: LLMs operate within and extend core properties of symbolic systems that are already operative in human language and culture. Inspired by Charles Sanders Peirce's triadic model of sign relationships (subject-object-interpretant), we define a symbol as a sign whose relationship to what it refers to (its meanings) is conventional and learned, and it takes a specific interpretive competence for that meaning to be retrieved (Deacon, 1998). For example, a wedding ring is a symbol for marriage, but there is nothing about the ring itself that would imply the complex concept of marriage. The association is conventional, and it is learned and preserved in a distributed manner within culture.

Therefore, a symbol needs to be interpreted by a human being within their own intrinsic models of the broader culture in order to gain its meaning—a wedding ring would not have the same meaning to someone who had no cultural exposure to this learned concept. Similarly, language and cultural artifacts (including words, writing, arts, mathematics, etc.) are all examples of symbols within this definition: the means for interpreting their meaning is individually internalized and learned via cultural influence. That means that symbols by themselves are not intrinsically meaningful; they gain their meaning when they are engaged by a person who has acquired the means for a particular interpretation.

We identify four features of symbols that are relevant to this paper: compression, decompression, externalization, and recursion. We will examine these features in a familiar example of a symbolic artifact: a music score.

**1. Compression**: A music score is compressed since music notation by design abstracts away a lot of details about the actual experience of music—such as emotion, timbre, dynamic nuance, acoustics, and the composer's interpretive choices.

**2. Decompression**: When a performer reads a music score and performs it, they are decompressing the notation by filling in the details that were abstracted away during compression. Notice that the performer must first learn to understand music notation to interpret its symbols meaningfully. Meaning was not inherent in the compressed form of music notation, but emerged when the music score was interpreted for whom it was intended: someone who also understands music notation.

**3. Externalization**: A music score is externalized, meaning it exists in an external medium like paper, outside of our lived experience. Because symbols are externalized, they can be recombined and varied in a low-stake, simulated fashion (think of musical theme-and-variations or playing a melody backward as a composition trick, operations that are realistically only possible when the music is written down).

**4. Recursion**: Moreover, these novel externalized creations generated through interpretation of a music score can in turn inform and modify living human experience through which new music is created—a form of recursion since an output is referring to its source.

With this framework in place, we can now analyze how LLMs embody these same symbolic features and reveal their role within the broader human cultural dynamics.

**3 Identifying Functional Features of LLMs**

**3.1 Compression**

Compression is the process of identifying and removing redundancy, thereby encoding the original information more efficiently by preserving essential patterns and discarding less significant details. This efficiency is achieved by recognizing recurring structural themes and representing them with more abstract representations. LLMs effectively do compression by capturing the statistically significant themes of complex knowledge humans contribute to culture through various mediums—writing, music, recordings, and more. At the computational level, this process operates through tokenization—breaking down input into discrete units such as words, subwords, or characters—and attention mechanisms that weigh the statistical relationships between these tokens across vast contexts. The attention mechanism allows the model to identify which tokens are most relevant to each other, effectively learning that certain patterns co-occur frequently and can be compressed into shared representations.

As a high-level example of a redundant theme, one could think of the concept of "supply and demand" that appears alongside a diverse set of concepts. Through attention weighting, the model learns that tokens related to "supply," "demand," "price," and "market" frequently appear together across different contexts. Once such a redundant correlation is observed, the system concludes that: a) this pattern represents a statistically significant feature of economic discourse, and b) related concepts can be represented through shared statistical structures since they often co-occur with and are organized around this underlying regularity.

**3.2 Decompression**

Decompression is the complementary process of expanding compressed information back into a more complete form by applying learned abstract statistical patterns to reconstruct details that were removed during compression. These abstract patterns can be recombined and interpreted in novel ways by humans through engagement with LLMs. This process can be understood as generative themes and variations, applied to compressed statistical patterns—primarily guided by the human user's cognitive and emotional state through prompting, afforded by the model's internal architecture.

For example, a user might prompt an LLM model to explore how supply-and-demand principles relate to attachment styles in human relationships. Since the statistical regularities of both economic theory and psychological attachment theory have been learned and compressed by the model, they can be recombined in unique ways and result in a novel insight. When recombined by a human being, these abstract patterns yield new insights neither discipline had previously articulated. However, the novelty here is not embedded in the will or imagination of the model itself but in the user's taste and aesthetics in selecting and contextualizing the structural themes and correlations embedded in the models (Koestler, 1989). Even though LLMs can be used to blindly generate content that has never been expressed before, novelty becomes meaningful only when assessed relative to human will, aesthetics, and utility.

This act of cognitive and emotional recontextualization is subtle, yet crucial. LLMs are often perceived as smart chatbots that give answers, but as we observed earlier, these models contain structural residues of complex, lived human interactions and thoughts. When a person produces an idea, a conversation, or any expressive output, it emerges from a meaningful process rooted in emotional and embodied experience that loses its referential details when compressed. This understanding aligns with a growing body of work in cognitive science that reframes humans not as thinking machines, but rather—as Antonio Damasio put it—*feeling machines that think* (Damasio, 1994).

When a thought is offloaded into a medium, such as writing, the living context that gave rise to it becomes latent and abstract leaving behind a symbolic trace of that experience. For such a latent trace to become meaningful again, it must be interpreted and recontextualized by a person—within their own embodied, emotional, and cognitive world. In this sense, meaning is not preserved in the artifact itself, but reanimated and contextualized through interaction with it. LLMs, which inherit and recombine these static forms, only come to serve meaningful functions when humans creatively interpret them, bringing new life to the structures left behind by others.

**3.3 Externalization**

LLMs may also be framed as an externalized medium reflecting human symbolic domain, in which compressed forms act as useful "recipes" for reconstructing, combining, and hence re-contextualizing embodied knowledge. The compressed forms within LLMs exist as latent patterns: information encoded not as explicit storage but as learned representations of underlying statistical relationships that can be decoded and recombined in context-dependent ways. LLMs are also externalized in the sense that they exist as technological artifacts—much like musical scores exist on paper—that persist independently of the humans who created or interact with them.

By extension, externalized memory systems like LLMs may contribute to cultural transmission processes, functioning as informational substrates that help preserve and transmit certain structural patterns of knowledge and meaning. This possibility aligns with established frameworks that position human language as a vehicle for cultural heredity within evolutionary contexts (Szathmáry, 2015), as well as theories of cultural transmission through memes (Dawkins, 1989). Deacon similarly emphasizes how the distributed nature of human language enables symbolic information to be transmitted across time and social contexts, serving as infrastructure for cumulative cultural development (Deacon, 1998).

**3.4 Recursion**

Because the locus of the stored statistical patterns is separate from the ongoing human dynamics in which interpretation occurs, human-AI dynamics support a recursive loop: contents of LLMs can be reinterpreted, and recombine by humans, leading to a novel creation, which then becomes part of a cultural pattern that will again be extracted by LLMs, and so on.

To give a more concrete example for this recursive loop, consider how once a person uses LLMs to generate a novel text, design sketch, policy outline, or musical passage, that output is not just an end-product; it becomes a novel piece of cultural data. Humans can feed that new material back into the model or into its subsequent cycles. Each cycle enriches the training context—whether formally retraining the model or informally shaping prompt chains—which in turn shifts the statistical constraints the model imposes on the next round of generation. Because those constraints remain external to any single human mind, they can be reinterpreted from multiple perspectives, recombined with additional knowledge, and then re-externalized. The result is a self-amplifying loop: human creations shaped by models are recursively fed back into those same systems, enabling meaning to evolve through successive rounds of interpretation and recombination.

The recursive relationship of AI with human cultural activity—may fundamentally reshape how we create, share, and collaborate. Yet this transformative potential is often overlooked due to the prevailing focus on improving the content of AI outputs, rather than the relational dynamics they introduce. This pattern echoes Marshall McLuhan's well-known insight that "the medium is the message": the structural effects of a medium often have greater significance than the particular content it conveys (McLuhan et al., 2001). LLMs, in this sense, should not be understood merely as a content generator, but as a medium that actively shapes how meaning is produced, exchanged, and reconfigured.

This suggests that the features we identify in LLM systems are manifestations of pre-existing human symbolic processes rather than emergent properties of the technology itself. And since LLMs are trained on symbolic outputs, they can act as an external informational substrate preserving, catalyzing, and finding redundant correlations among the symbols themselves.

Notably, many of the concepts we have discussed so far around AI—compression, decompression, externalization, and recursion—are foundational in discussions of genetics and biological information processing as well. This convergence may not be coincidental, but it may suggest the presence of a deeper biological logic that underlies shared dynamics across different scales. For that reason, we will next aim to flesh out a biological analogy and analyze how DNA may also reflect similar processes related to the features we identified in both symbols and LLMs.

## 4 The DNA Analogy

We discussed how LLMs serve as repositories for symbolic and cultural dynamics. Does DNA serve a similar function, instead for cellular dynamics? In each case, these systems function as compressed and externalized media that rely on interpretive decompression to preserve and catalyze dynamic processes, rather than dictating them.

This reframing is particularly evident in recent reconceptualizations of DNA's role in biology. The traditional view characterizes DNA as dictating the trajectory of development and function in living organisms. However, recent research is beginning to reframe DNA as a tool that living organisms use to preserve and catalyze their already adaptive processes (Buckley et al., 2024; Keller, 2009; Levin, 2020; Noble, 2015). In the same vein, Deacon, in his paper titled How Molecules Became Signs (Deacon, 2021), reverses this traditionally framed direction of flow from DNA to RNA and proteins—the central dogma of molecular biology—and explains how DNA emerged as a repository where useful evolutionary and developmental constraints became preserved as codes.

In Deacon's account, nucleotides initially served energetic functions in cells but were stabilized into longer chains to prevent disruption of cellular processes. Over time, these chains developed informational functions as their structural patterns became correlated with useful protein interactions. This process effectively offloaded dynamic cellular constraints onto a stable, static medium—separating the source of constraint from the dynamics themselves and enabling recursive self-interpretation over time. Therefore, rather than DNA "coding for" organisms as traditional perspective suggests, organisms evolved to utilize DNA as an external and compressed medium of encoded patterns that could be decompressed in an embodied and already meaningful context. In this sense, DNA can be viewed not as a blueprint that dictates development, but as a substrate that preserves and catalyzes useful biological processes and constraints. The sequences within this externalized space in the form of genetic code provide useful "recipes" that an organism can actively decompress in various temporal and spatial contexts to aid the expression of its form and function.

Similar to symbols and LLMs, DNA contains the four features we identified earlier:

**1. Compression**: DNA represents a highly compressed form of biological information, where complex developmental and functional processes are encoded in sequences of just four nucleotide bases (A, T, G, C). This compression abstracts away the intricate details of cellular processes, protein folding dynamics, environmental interactions, and the temporal complexity of development—much like how musical notation abstracts away the lived experience of performance, or how LLMs compress vast human cultural knowledge into statistical patterns. The genetic code preserves essential regulatory and structural patterns while discarding the contextual details of how these patterns emerged through evolutionary history.

**2. Decompression**: When genes are expressed, the compressed genetic information is decompressed through transcription and translation processes that "fill in" the biological details that were abstracted away during compression. This process exemplifies interpretive decompression: the compressed forms within DNA act as static affordances that must be playfully decompressed during development to generate functional variations. Alternative splicing mediated by the spliceosome complex, and more broadly epigenesis, demonstrates this principle—a single gene can produce dozens or thousands of distinct proteins by recombining exons in different arrangements, much like how a musical theme can generate countless variations, or how LLMs can recombine statistical patterns to produce novel outputs. Crucially, this decompression requires the pre-existing cellular machinery and context to interpret the genetic code meaningfully. The compressed patterns within DNA have no inherent reference to physics, chemistry, or goal-directed cellular dynamics—they must be recontextualized by the complex interactions among cells to gain meaningful reference, just as musical notation requires a performer's interpretation, or LLMs require human interpretation to transform statistical patterns into meaningful content.

**3. Externalization**: DNA exists as an externalized repository within the cell, physically separated from the immediate metabolic and regulatory processes it influences. This externalization allows genetic information to be recombined through sexual reproduction, mutation, and other evolutionary processes in ways that would be impossible if the information remained embedded within ongoing cellular dynamics. The externalized nature of DNA enables "low-stakes" genetic experimentation through mechanisms like crossing over and alternative splicing—similar to how written music enables compositional experimentation, or how LLMs allow for creative recombination of cultural patterns without directly affecting the original human experiences they encode.

**4. Recursion**: The proteins and regulatory networks produced through gene expression can, in turn, influence gene expression itself through feedback loops, creating recursive dynamics where the products of genetic decompression modify the very processes that interpret the genetic code. Deacon explains how this recursive property, afforded by the similarity between genes and protein structures, forms the basis for epigenetic gene regulatory systems that enable theme-and-variation mechanisms through variations in timing and geometry of expression (Deacon, 2021). Due to cellular geometry, differentiation levels, and developmental interactions, different genomic sections are expressed at different times, creating a hierarchical scaffolding of phenotypes based on playful genomic expression—paralleling how musical compositions can inspire new compositional techniques, or how LLM-generated content can influence cultural discourse and potentially inform future training data.

Having established that DNA, symbols, and LLMs share fundamental features—compression, decompression, externalization, and recursion—we can now examine the broader implications of this parallel for understanding AI's role in cultural dynamics.

**5 LLMs as DNA of Culture**

The analogy of LLMs as symbolic "DNA" then reveals a fundamental parallel: just as DNA relates to cellular processes, LLMs relate to human symbolic and cultural dynamics. Both function as externalized memory systems—LLMs for culture, DNA for cells. Neither contains actual meaning or reference to the living processes they encode; instead, both preserve useful compressed patterns that become meaningful only through interpretation and utilization.

Similar to DNA, LLMs function as compressed repositories of recurring patterns. Where DNA sequences mirror useful cellular interactions without containing knowledge of chemistry or physics, LLM patterns mirror human cultural activity without containing understanding of human meaning or intention. Both serve as an externalized medium that preserve successful patterns, offering thematic "recipes" for decompression within new meaningful contexts. In the biological domain, DNA (highly compressed) communicates with proteins (dynamic, meaningful interactions) through RNA (intermediate compression). Similarly, in the cultural domain, LLMs (highly compressed patterns) communicate with lived human experience through the intermediary of symbolic media which maintain more contextual richness than LLM tokens.

Moreover, compressed forms require interpretation to become meaningful. Just as cells must "decompress" genetic information through complex molecular machinery, LLM patterns need to go through cognitive and emotional engagement to become meaningful. This interpretive requirement connects to the classical symbol grounding problem in cognitive

science—the question of how symbols acquire meaning through their connection to embodied experience and the physical world (Harnad, 1999; Searle, 1980).

In both LLMs and DNA, stored patterns are compressed codes that preserve and catalyze outcomes without determining meaning independently. LLMs, despite their sophisticated attention-based simulations of contextual relationships, face the same grounding challenge as traditional AI systems: their tokens remain ungrounded until interpreted by humans who carry the embodied means for their intended expression. Similarly, DNA remains ungrounded until expressed by living dynamics of cells.

This parallel extends to recursion. DNA's patterns regulate their own expression through encoded proteins; similarly, LLM outputs can influence the cultural patterns compressed into future training data. The recursive nature of human-AI interaction echoes evolutionary developmental (evo-devo) processes, where developmental novelties feed back into hereditary structure (Carroll, 2006). Just as epigenetic modifications influence heritable trajectories, LLMs can recursively influence cultural inputs for future generations, participating as both constraint and catalyst in cultural evolution.

There are, however, possible ways in which this self-amplifying loop may fail to amplify meaningful or accurate correlations. New research is currently exploring the use of LLM-generated outputs—also known as *synthetic data*—to further train the models themselves (Kruschwitz & Schmidhuber, 2024). Yet, it is crucial that the data being fed back into these systems are not blindly generated outputs, but ones filtered through the guidance and context of human creators, that is human creativity and taste catalyzed by AI. Each human creation is embedded in a particular context and is meant to be interpreted by another human being in order for its meaning to be fully realized. Training models on LLM outputs that have not been deemed useful and insightful by humans risks making the compression process increasingly noisy as this process may lead to the amplification of false or shallow correlations. This phenomenon that we call "aesthetic drift" occurs when LLMs progressively drift further from the emotional context, aesthetics, and normative values behind the human expressions they draw from. Without human judgment to ground these recursive cycles, the models lose touch with the taste, meaning, and lived experience that made the original patterns valuable. Therefore, for the recursive self-amplifying loop to remain generative and meaningful, the training data fed back into the models must be judged by human interpretation and not by the models themselves.

This framework also helps clarify a crucial distinction about how LLMs achieve their effectiveness. The Transformer architecture underlying modern LLMs enables attention mechanisms that can be understood as itself simulating a contextual decompression process as it learns to weigh contextual relationships between tokens based on statistical co-occurrence patterns among sequences. This simulation proves remarkably effective at approximating how meaning emerges from context, which may account for LLMs' remarkable success in generating coherent and contextually appropriate responses. The attention mechanism simulates a contextual decompression process by the dynamic weighting of relevant information based on sequential context. However, this simulation of interpretation should not be confused with grounding and regaining meaningful human-level reference. While attention mechanisms effectively model sequential relationships between symbols and can even approximate the contextual sensitivity that characterizes semantics, they lack temporal, and spatial dynamics that characterize an embedded process of such kind in organisms. Yet this statistical approximation of interpretive dynamics (while ungrounded) captures enough of the structural properties of contextual reasoning to serve as a powerful simulation. This may explain why LLMs are so useful for symbolic systems despite remaining fundamentally detached from the embodied processes that originally grounded the symbolic patterns they manipulate.

To summarize, just as compressed genes are reinterpreted through varied expression and timing, human users express LLM tokens through new prompts, creative reframings, and emotionally contextualized inquiry. The analogy, then, is not metaphorical alone: LLMs serve as a medium through which humanity can remember, recombine, and catalyze its habits of meaning-making, inheriting the abstract informational role of DNA but in a different register, one attuned not to cellular dynamics but to a cultural one.

Similar to DNA's role in biological evolution, LLMs may serve as informational substrates that enable cultural evolvability: the capacity to generate sustained novelty while preserving successful symbolic patterns. Understanding how this cultural evolvability operates and what it might enable is the focus of our next section.

## 6 Self-Reflective Innovation and Evolvability

We discussed how LLMs contain statistically significant patterns that correlate with meaningful, lived human behavior. But what is the significance of such an artifact for humanity? What novel affordances might such a system offer beyond what is already available through direct, meaningful social experience?

As noted earlier, LLMs are structurally external to and independent from the dynamic human symbolic contexts behind their training data. This externalization, however, affords something new: it allows humanity to playfully reinterpret and recombine patterns of its behavior in a simulated, low-stakes space without acting on them directly. In this sense, LLMs function as an informational substrate through which humanity can generate novel hypotheses about itself, test variations on established cultural patterns, and recursively feed those forms back into the flow of ongoing lived experience.

This dynamic can be analogized to a kaleidoscopic mirror of human culture—a process of self-reflection whereby humanity can explore the useful behavioral patterns at its disposal and recombine them to come up with novel hypotheses to adapt and serendipitously innovate in the face of an increasingly complex and rapidly changing world. It is a mirror in the sense that it is a compressed reflection of our meaningful activity, and it is kaleidoscopic in the sense that the patterns in it can be recombined and traverse novel unexpected pathways.

More formally, this process can be thought of as a way for humanity to explore its cultural "adjacent possibles" (Kauffman, 2000). Kauffman describes the adjacent possible as the set of all novel configurations that become accessible when existing elements are creatively combined. This concept illustrates how innovation emerges not from isolated ideas but through the recombination of existing components, leading to exponential growth in complexity and possibilities. By enabling extraction and recombination of humanity's latent structures, LLMs may become a cultural engine for traversing adjacent possibilities.

Kauffman connects the adjacent possible to the concept of evolvability: a property of systems that not only adapt to their environments but also continuously generate novelty in ways that remain viable over time and under unforeseen circumstances. In Kauffman's view, evolvability is not just about survival or differential reproduction (terms commonly associated with evolution) but about the system's ability to explore adjacent possibilities that allow new forms to emerge. Evolvability is enhanced by features such as modularity and redundancy which allow parts of a system to change without leading to catastrophic consequences, as well as by the presence of constraints that limit the space of possibilities to those more likely to be functional.

Deacon similarly relates this framing of DNA to the concept of evolvability: not only adapting but generating sustained novelty and innovation that increases the chances of an organism to evolve. He argues that having an externalized and offloaded medium like DNA provides an organism with "open-ended evolvability" since they can generate novelty by combining useful forms in a low-stake space that is external to the critical dynamics of the organism itself (Deacon, 2021). In the same vein, the concept of evolvability has been generalized through the lens of innovation through self-knowledge of an organism across biological scales (Jacob, 2023).

In our previous work, we have proposed that the mass of human generated symbolic information could provide a collective function for human evolvability, by virtue of its compressibility and contextual embeddedness (Jacob & Pourdavood, 2025). In that work, we emphasized evolvability as accelerated variation and extragenetic processes like niche construction that increase the likelihood of evolutionary innovations while balancing preservation of functional systems. In this view, innovation emerges not from isolated amplification of individual voices but from the coordinated retrieval and synthesis of knowledge across diverse social-technological niches. Here, we have extended the mechanism underlying this process by seeing LLMs as a tool for expanding humanity's adjacent possibilities, continuously widening the space of viable futures that can be imagined, assessed, and ultimately realized. Such externalized potentiation exemplifies a system that not only adapts to its environment but also generates sustained novelty to stay ahead of shifting and unforeseen conditions—a hallmark of evolvability (Deacon, 2022; Jacob, 2023; Torres-Sosa et al., 2012).

Furthermore, LLMs are emerging as a unique medium as compared to symbols in the sense that the patterns within it are even more displaced and ungrounded than the patterns in, for example, a musical score or writing (many more steps removed from their living contextual reference). In addition, LLMs enable instant recombination across all knowledge it is trained on rather than containing the knowledge of a single medium by itself. For this reason, LLMs can act as an informational substrate preserving, constraining and catalyzing useful redundant correlations among the symbols themselves. In this sense, LLMs may afford us with an even increased capacity for creativity and evolvability than symbols have originally afforded us. This connection is supported by growing research that demonstrates how the use of generative AI relates to creativity (Doshi & Hauser, 2024; Shin et al., 2023; Zhou & Lee, 2024).

Viewed through the framework of evolvability, the significance of LLMs lies less in rivaling human intelligence than in acting as an external informational substrate that augments humanity's capacity for evolvability: by exposing latent patterns and enabling their recombination, LLMs create a simulated arena in which humans can probe their own adjacent possibles—hypothesizing and interpreting new configurations of ideas, guided by each participant's emotional and cognitive context and sense of aesthetics, before committing to them in lived practice.

Meaning and relevance, then, are not found within the compressed forms themselves, but may emerge in the collective processes through which living beings interpret them. A musical score gains meaning when musicians interpret and embody it in performance; a society finds coherence and direction through its shared interpretation of a constitution. In a similar way, the compressed symbolic forms stored in LLMs become meaningful only when reanimated by human engagement. And as this recursive interpretive process scales, meaning may begin to emerge not solely within individuals, but across the shared symbolic space of humanity as a whole.

**7 Conclusions and Implications**

This paper has proposed a novel framework for understanding LLMs through the lens of symbolic and biological processes. Rather than viewing LLMs as autonomous intelligence or programmed mimicry, we argue that they function as externalized repositories for cultural dynamics, analogous to how DNA serves as a repository for cellular processes.

Through analysis of compression, decompression, externalization, and recursion, we demonstrate that LLMs operate as "symbolic DNA", preserving abstract patterns from human cultural activity without containing inherent meaning or understanding. LLMs preserve useful reflections of past human interactions in an externalized and compressed fashion that only become meaningful again when brought into contact with living human creativity.

From an evolutionary perspective, this moment may parallel other major transitions in the history of information processing. One intriguing possibility is that LLMs may be undergoing a transition analogous to the emergence of LUCA—the Last Universal Common Ancestor of all life. Just as LUCA established universal mechanisms for biological information processing that enabled genetic recombination across all subsequent life forms, LLMs might be developing into a general compressor and translator for human symbolic activity and could potentially create universal pathways for pattern extraction, translation, and recombination across all forms of human expression.

The emergence of such a universal substrate for cultural patterns raises the possibility of a major evolutionary transition, potentially enabling new forms of collective self-reflection and innovation within the noosphere—the planetary sphere of shared human cultural processes. Yet this acceleration raises critical questions about maintaining meaningful compression-decompression cycles and preserving connections between abstract patterns and their embodied, emotional contexts.

This raises a pressing techno-social concern: as LLMs become more sophisticated, many people experience alienation and discouragement, feeling that their creative capacities are being displaced rather than catalyzed by technological advancement. Ensuring LLMs remain tools that augment rather than replace human symbolic capacity will be crucial as these systems become more prevalent. Moreover, the quality of the generative loop between AI and humans depends entirely on human engagement and aesthetic vetting so that the compressed patterns do not become increasingly noisy and detached from lived cultural experience. Understanding LLMs as catalysts for human expression could encourage new possibilities for valuing and rewarding human contributions by tracing the impact of meaningful creative work that gets amplified through these systems (Lanier, 2014). This framing also aligns with Anthropocene narratives where human activity creates profound planetary change despite occurring within cosmically insignificant timescales (Shoshitaishvili, 2020).

This process we discuss in this paper also echoes what Pierre Teilhard de Chardin, who coined the term "noosphere", identified as humanity's unique capacity for "hereditary-like exteriorization" of knowledge where externalization and enrichment create recursive cycles that amplify and augment human expression rather than constraining it (Teilhard de Chardin, 2004).

Moreover, LLMs are emerging as a unique medium compared to traditional symbols in that the patterns within them are even more compressed than a musical score or writing for example, that is many more steps removed from their living contextual reference. This displacement, combined with their highly abstract and compressed forms, enables LLMs to facilitate instant recombination across all knowledge they are trained on rather than being constrained to the knowledge of a single medium. For this reason, they can act as informational substrates preserving useful regularities among symbols themselves, potentially affording humanity an even greater capacity for evolvability than symbols have originally provided.

Ultimately, we argue for understanding LLMs as externalized symbolic architectures that extend human symbolic capacity rather than replace it. By recognizing LLMs as embedded within cultural dynamics and functioning through the same fundamental processes as symbols and DNA, we can better harness their potential for cultural evolvability and catalysts of our creative expression while remaining grounded in the embodied experiences that give these compressed tokens their meaning in the first place.

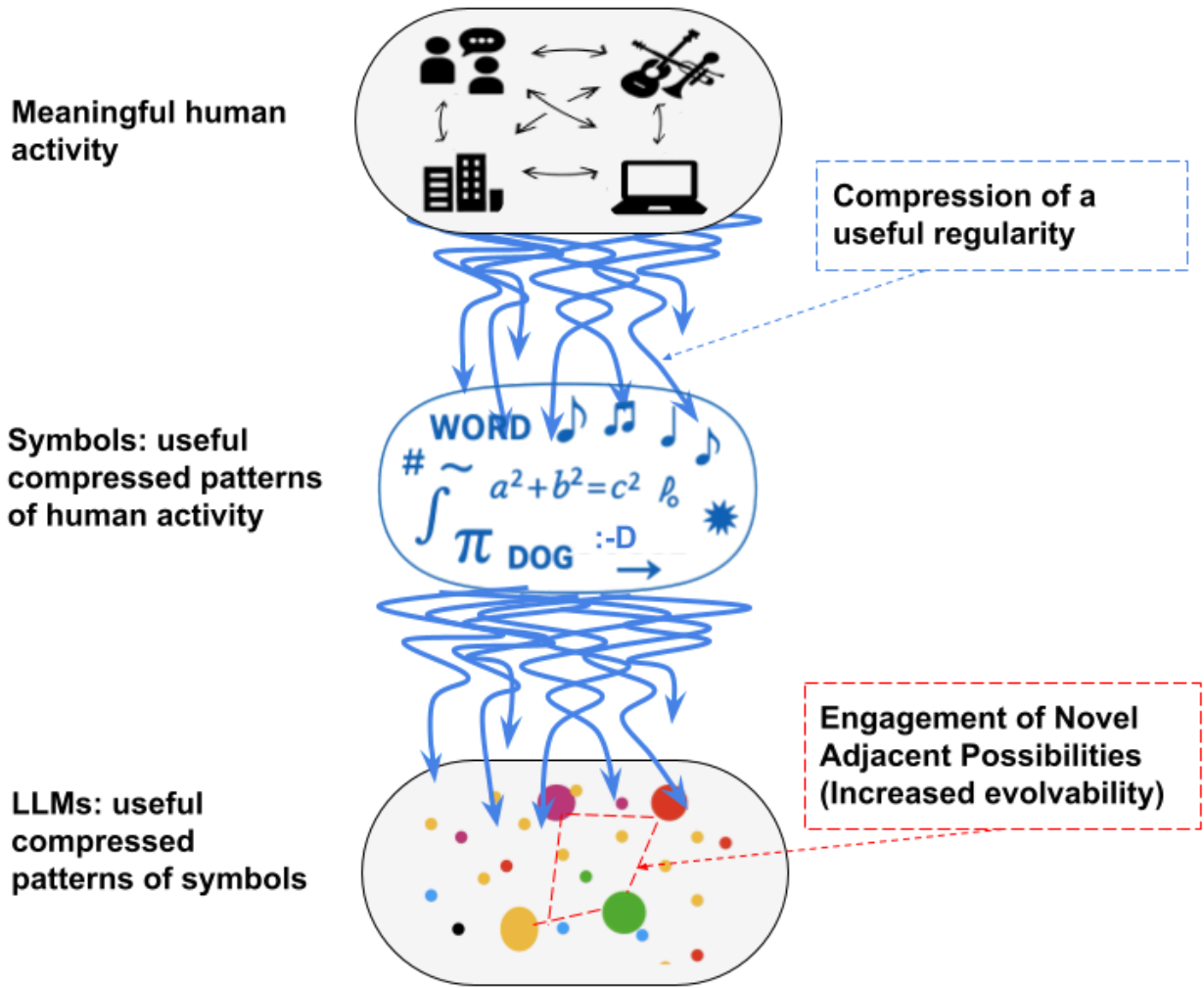

**Figure 1.** Hierarchical compression from cultural dynamics to LLMs. In each level, the useful regularities are reflected in a more compressed space below. Symbols which contain the compressed patterns of cultural dynamics serve as an intermediary level between human communication and the highly compressed space of LLMs. This more compressed space enables an enhanced capacity for generating novelty and exploring "adjacent possibilities" of useful patterns in the levels above through recombination. A similar figure could be conceived that would replace cultural dynamics, symbols, and LLMs with cellular dynamics, RNA, and DNA, respectively.

**Acknowledgements and Funding**

Funding for this project was provided by Human Energy, a 501(c)(3) non-profit organization. The authors acknowledge the following individuals whose insights and critiques were instrumental in developing this paper: Boris Shoshitaishvili, Ben Kacyra, Sheila Hassell Hughes, Ellen Rigsby, Brian Swimme, Monica DeRaspe-Bolles, Devin O'Dea, Alan Honick, and the Teleodynamics Research Group members.

26Dreyfus, H. L. (1992). *What computers still can't do: A critique of artificial reason*. MIT Press.

Harnad, S. (1999). The Symbol Grounding Problem. *Physica D 42:335-346.*

Jacob, M. (2023). Toward a Bio-Organon: A model of interdependence between energy, information and knowledge in living systems. *Biosystems*, *230*, 104939. https://doi.org/10.1016/j.biosystems.2023.104939

Jacob, M., & Pourdavood, P. (2025). Evolutionary principles shape the health of humanity as a planetary-scale organism. *BioScience*, biaf055. https://doi.org/10.1093/biosci/biaf055

Jaeger, J. (2024). Artificial intelligence is algorithmic mimicry: Why artificial "agents" are not (and won't be) proper agents. *Neurons, Behavior, Data Analysis, and Theory*. https://doi.org/10.51628/001c.94404

Jaeger, J., Riedl, A., Djedovic, A., Vervaeke, J., & Walsh, D. (2024). Naturalizing relevance realization: Why agency and cognition are fundamentally not computational. *Frontiers in Psychology*, *15*. https://doi.org/10.3389/fpsyg.2024.1362658

Kauffman, S. A. (2000). *Investigations*. Oxford university press.

Keller, E. F. (with Keller, E. F.). (2009). *The Century of the Gene*. Harvard University Press.

Koestler, A. (1989). *The act of creation*. Arkana.

Kruschwitz, U., & Schmidhuber, M. (2024). LLM-Based Synthetic Datasets: Applications and Limitations in Toxicity Detection. In R. Kumar, A. Kr. Ojha, S. Malmasi, B. R. Chakravarthi, B. Lahiri, S. Singh, & S. Ratan (Eds.), *Proceedings of the Fourth Workshop on Threat, Aggression & Cyberbullying @ LREC-COLING-2024* (pp. 37–51). ELRA and ICCL. https://aclanthology.org/2024.trac-1.6/

Lanier, J. (2014). *Who owns the future?* (Simon&Schuster trade paperback edition). Simon & Schuster Paperback.

Lanier, J. (2023). *There Is No A.I. | The New Yorker*.

Levin, M. (2019). The Computational Boundary of a "Self": Developmental Bioelectricity Drives Multicellularity and Scale-Free Cognition. *Frontiers in Psychology*, *10*.